\documentclass[lettersize,journal]{IEEEtran}
\usepackage{amsmath,amsfonts}
\usepackage{amssymb}
\usepackage{algorithmic}
\usepackage{algorithm}
\usepackage{array}
\usepackage{multirow}
\usepackage{booktabs}
\usepackage{tabularx}
\usepackage{textcomp}
\usepackage{stfloats}
\usepackage{url}
\usepackage{verbatim}
\usepackage{graphicx}
\usepackage{cite}
\usepackage{dsfont}
\usepackage[commandnameprefix=ifneeded]{changes}
\usepackage{subcaption}
\begin{document}

\title{Revisiting data-driven dynamic security assessment with a tabular foundation model}


\author{\IEEEauthorblockN{Olayiwola Arowolo$^{a}$, Maosheng Yang$^{a}$, Jochen L. Cremer$^{a}$$^{,b}$}\\
\IEEEauthorblockA{$^{a}$Delft University of Technology, Delft, the Netherlands \\
$^{b}$Austrian Institute of Technology, Vienna, Austria \\
o.a.arowolo, m.yang-2, j.l.cremer\{@tudelft.nl\}\\
}
}

\maketitle

\begin{abstract}
Data-driven pre-fault dynamic security assessment (DSA) rapidly evaluates the dynamic risk of credible contingencies on a power system using machine learning (ML). Existing approaches face two limitations. First, they require a large labelled database for training, with a separate model trained, tuned, and maintained for each contingency in a potentially long list of credible contingencies. Second, the trained models generalize poorly to unseen contingencies. This work addresses these limitations by using a tabular foundation model (TFM) that assesses stability through in-context learning, and requires no retraining or hyperparameter optimization. 
We use a single TFM to assess many contingencies at once, removing the need to train one classifier per contingency.
We also characterize when using electrical distance coordinates (EDC) as continuous features enables the TFM to generalize to unseen contingencies and when it does not, showing how a few labelled samples can reliably improve generalization. 
Through comprehensive case studies on the IEEE 68-bus system, we show that a single TFM attains an average Macro F1 score of about 90\% with only 120 labelled samples per contingency, outperforming both classical ML and deep learning baselines, without any model retraining or hyperparameter tuning. For held-out contingencies, we show that using just 10 labelled samples of the new contingency with a modified EDC encoding matches the balanced accuracy of the best achievable transfer learning oracle model, which requires fully labelled data and is not deployable in practice. 
Overall, this initial study paves the way for developing and deploying foundation models for power system operations, with possible applications across multiple operational tasks.
\end{abstract}

\begin{IEEEkeywords}
Dynamic security assessment, machine learning, tabular foundation model, electrical distance coordinate, generalization.
\end{IEEEkeywords}

\section{Introduction}
\IEEEPARstart{T}{he} increasing penetration of renewable energy increases variability and uncertainty in power grid operations \cite{panciatici_operating_2012}. Despite these changes, the power grid must remain reliable and secure, as equipment failures can lead to cascading outages with disastrous consequences \cite{zhang_mitigating_2024}. Operators have to balance pushing the power system near its limit for economic efficiency and maintaining secure operations \cite{bellizio_machine-learned_2022}. Dynamic security assessment, which involves evaluating the ability of the power system to withstand a sudden contingency, enables the power system to be operated less conservatively, without sacrificing system security \cite{panciatici_operating_2012}.

Dynamic security assessment involves solving differential-algebraic equations at small time steps to create time-domain simulations (TDS). These simulations can be computationally intensive, especially for large grids. Thus, the use of TDS to assess system stability has typically been limited to preventive screening, as simulations are too time-consuming for online use \cite{duchesne_recent_2020}. The work in \cite{wehenkel_automatic_1998} introduced a data-driven approach for online DSA. The approach is based on training a machine learning (ML) classifier model with historical or TDS-generated data during an offline phase. Once trained, the model can be used to rapidly assess the system's stability during an online phase \cite{savulescu_real-time_2014}. Classic ML techniques such as decision trees \cite{bugaje_selecting_2021} and support vector machines (SVM) \cite{tang_power_2018} are commonly used in the literature because of their fast assessment and good interpretability \cite{song_steady-state_2025}. We refer the reader to \cite{duchesne_recent_2020} for a comprehensive overview of classic ML methods employed in DSA.

The advent of deep learning has seen the application of deep neural networks (DNNs) to online DSA. DNNs can extract features directly from PMU data streams to predict a stability label or the system's response trajectory \cite{yu_intelligent_2018, gupta_online_2019}.  The use of DNNs for DSA has led to two classes of data-driven DSA approaches, namely, pre-fault DSA and post-fault DSA. Pre-fault DSA uses steady-state operating conditions (OCs) as features (voltages, power demands, etc.), which can be collected before any fault occurs. On the other hand, post-fault DSA uses time series trajectory measurements from PMUs, sampled before, during or immediately after fault clearing to predict the system's response \cite{ cui_frequency_2024, karacelebi_online_2025}. For example, physics-informed neural networks (PINNs) have been used to predict the response of the system to disturbances by approximating the post-fault state trajectories. The goal of pre-fault DSA is to assess as many contingencies as possible before they occur, and it usually runs minutes to hours in advance to enable preventive control actions by operators \cite{qi_continuous_2025}. On the other hand, post-fault DSA aims to activate automatic emergency controls \cite{karacelebi_online_2025}. This paper focuses on pre-fault DSA.

\begin{figure}[t]
    \centering
    \includegraphics[width=0.48\textwidth]{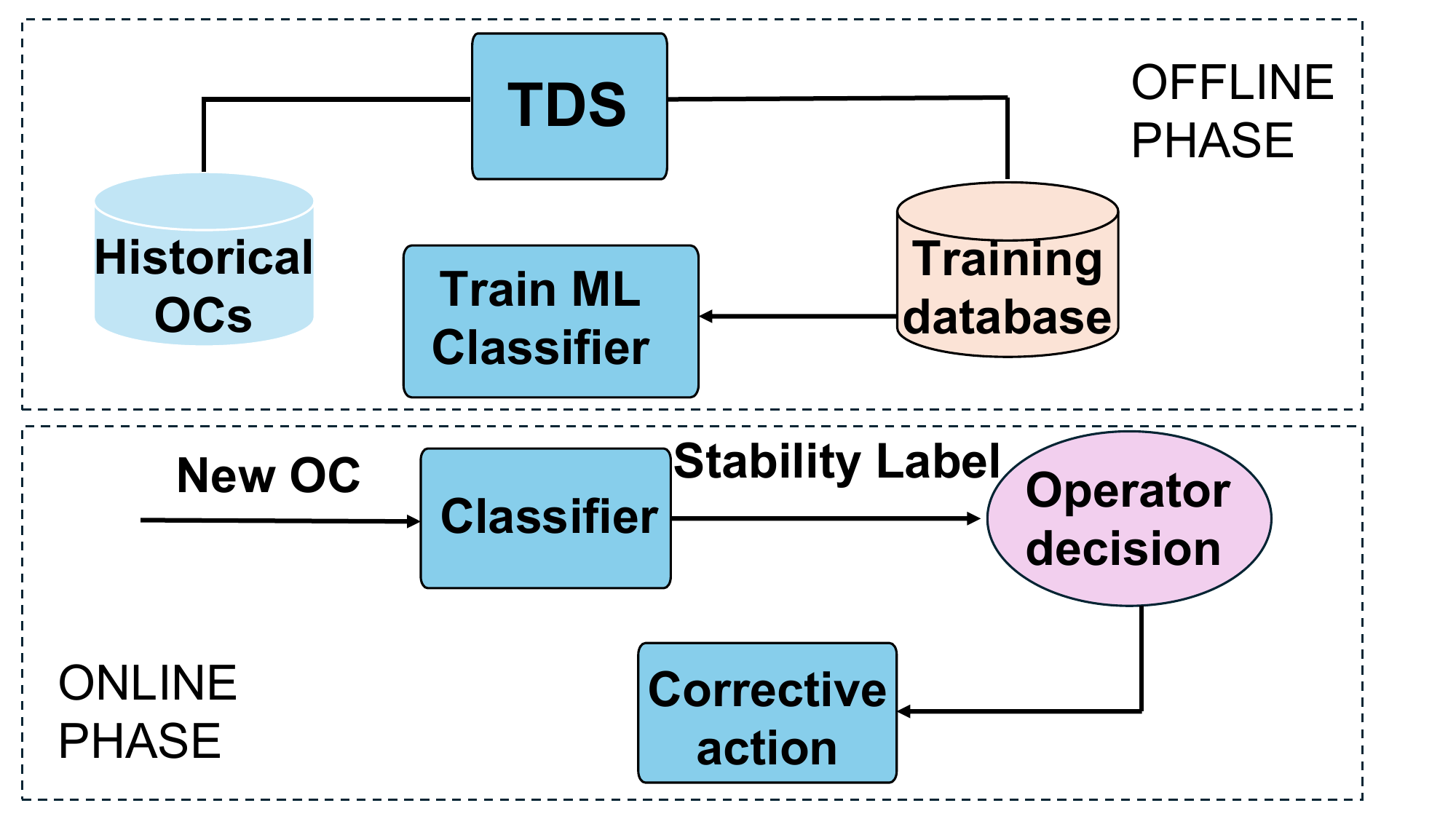}
    \vspace{-1mm}
    \caption{\footnotesize Conventional data-driven pre-fault DSA workflow. In the offline phase, a classifier is trained for every contingency to be assessed during the online phase.}
    \label{fig:DSA_workflow}
\end{figure}

Fig. \ref{fig:DSA_workflow} shows the typical workflow for an ML-based pre-fault DSA approach. A training database is constructed by running TDS for a distribution of realistic OCs. A classifier is trained for each fault in the list of credible contingencies \cite{konstantelos_implementation_2017}. It is assumed that the most severe contingencies are included in the contingency list, so that the DSA approach works for known contingencies. However, for unforeseen contingencies outside the expected contingency list, the classifier may have poor accuracy. While this approach is currently preferred by operators \cite{liu_bayesian_2020}, there may be unexpected contingencies which compromise the accuracy of the DSA system. Also, there can be a significant number of credible contingencies in a large power system, which makes the process of training and updating hundreds or thousands of individual classifiers cumbersome \cite{konstantelos_implementation_2017}.

There have been approaches to improve generalization to new/unforeseen contingencies in the literature. The work in \cite{ren_transfer_2020} proposed a transfer learning approach for adapting to a new but related contingency, while the same authors proposed the use of multi-label learning for training a classifier for many faults with missing/incomplete labels \cite{ren_pre-fault_2023}. These approaches do not extend to completely different or unrelated contingencies. The authors of \cite{qi_continuous_2025} proposed representing contingency locations with electrical distance-based coordinates (EDC) for better generalization. However, while contingencies at similar locations usually have similar DSA labels, pre-fault information alone is insufficient for characterizing the dynamic behaviour of the system. Other approaches for handling new contingencies include online learning \cite{sun_online_2007}, whereby the classifier is retrained periodically or whenever the accuracy or some other metric falls below a specified threshold \cite{bellizio_machine-learned_2022}.
Despite these innovations, the generalization of pre-fault DSA models to unforeseen contingencies which are not part of the training database or changing topologies remains challenging \cite{arowolo_exploring_2025}. Consequently, ineffective generalization to changing system topologies is one of the key challenges hindering the practical adoption of ML-based DSA systems. \cite{duchesne_recent_2020}.

Without effective generalization, the combinatorial explosion of sampling every feasible OC and topology combination makes it impractical to generate exhaustive data \cite{arowolo_exploring_2025}. Therefore, the computational budget for generating labels via TDS is a limiting constraint for data-driven pre-fault DSA. Since ML classifiers require fairly balanced classes, and insecure operating conditions occur only rarely under naive Monte Carlo sampling, a large number of simulations must be run to obtain a sufficient number of insecure cases to characterize the decision boundary. This leads to a very inefficient use of the limited computational budget for TDS. Hence, a key question is: how can the limited computational budget be best spent? While smarter data generation methods for pre-fault DSA databases have been proposed, including importance sampling \cite{liu_systematic_2014}, exploring stability boundaries \cite{giraud_dataset_2025}, and diversity sampling \cite{bugaje_split-based_2023}, they may also incur high computational cost due to optimization steps required to reduce the search space.

To reduce the computational burden of generating labelled datasets for multiple contingencies under pre-fault DSA, it is desirable to require minimal labelled data for each contingency considered and to generalize to unseen contingencies. This motivates the development of approaches that are sample-efficient, generalize across systems or faults, and remain robust to class imbalance. Foundation models are promising in this regard. Foundation models (FMs) are trained on large-scale, heterogeneous datasets and can be fine-tuned for different applications using small datasets. The idea of a foundation model for power systems has been proposed in \cite{hamann_foundation_2024}, with initial use cases which focus on steady-state operations. \cite{li_predicting_2026} uses a foundation model to predict power system response trajectories, within the scope of post-fault DSA. However, the application of foundation models to pre-fault DSA remains unexplored.

In this work, we present a new approach for pre-fault DSA based on a pretrained tabular foundation model (TFM). The model works on small-sized data samples, requiring no retraining or fine-tuning step. To further improve generalization of the TFM to unseen contingencies when a few labelled data are available, we modify EDC (mod-EDC) to capture post-fault topology. The main contributions of this work are as follows:

\begin{itemize}
    \item We explore the application of a TFM for pre-fault DSA. We show that the TFM enables accurate DSA prediction with a much smaller training dataset than currently assumed, without retraining or hyperparameter optimization. To the best of our knowledge, this is the first application of a TFM to pre-fault DSA.
    \item We propose a multi-contingency DSA workflow using a single TFM, which removes the need to train one separate classifier for each contingency. The proposed approach outperforms the multi-contingency DSA (MDSA-S) baseline  
    \item We characterize how EDC enables the TFM to generalize to unseen contingencies and when it fails. We further show that a simple modification of EDC (mod-EDC) improves the generalization only when a few labelled samples of the previously unseen contingency are available. 
\end{itemize}

The remainder of the paper is structured as follows: Section II formulates the pre-fault DSA problem. Section III presents the proposed workflow. In Section IV, we perform case studies to evaluate the proposed approach.  Section V discusses the results and notes some limitations, while Section VI concludes the paper and discusses future directions.

\section{Data-driven pre-fault DSA}

\subsection{Problem statement}
Power system dynamics can be described by a set of differential-algebraic equations
(DAEs):
\begin{equation}
\label{eq:dae}
\begin{cases}
\dot{x} = f(x, g, a), \\
0      = h(x, g, a),
\end{cases}
\end{equation}
where $x \in \mathbb{R}^{N}$ are the dynamic state variables, $g \in  \mathbb{R}^{M}$ are the algebraic states and $a$ are control states. Here, $f(\cdot)$ is a function which describes the evolution of the system states, while $h(\cdot)$ are the algebraic network constraints which the system must satisfy. 
Under changing generation and loading conditions, the system must be able to withstand a sudden contingency, such as a three-phase fault on a line or sudden loss of a large load. To determine the system's dynamic response, the DAEs in Eq. \ref{eq:dae} are solved through numerical integration at small time steps:
\begin{equation}
\label{eq:tds}
x_{n+1} = x_{n} + \int_{t_{n}}^{t_{n+1}} f(x, g, a)\, dt.
\end{equation}
This is known as a \emph{time-domain simulation} (TDS). After the TDS, the stability of the system can be determined, for example, by the Transient Stability Index (TSI):
\begin{equation}
\label{eq:tsi}
\mathrm{TSI} = \frac{360 - |\Delta \delta_{\max}|}{360 + |\Delta \delta_{\max}|},
\end{equation}
where $|\Delta \delta_{\max}|$ is the maximum power-angle difference between any two generators after a fault. A system with $\mathrm{TSI} > 0$ will be marked stable; otherwise, it is unstable. The stability label is denoted by $Y$. For a large system, $x$ may include thousands of variables, making real-time solution of the DAE challenging.
With data-driven pre-fault DSA, the TDS process described above is used to generate a training database. The database is then used to train a classifier model, which uses pre-fault steady-state features $X \in \{P,\, Q,\, V,\, \theta\}$ to predict an approximate label $\tilde{Y}$:
\begin{equation}
\label{eq:classifier}
\mathcal{\tilde{F}}(X) \rightarrow \tilde{Y},
\end{equation}
where $P,\, Q,\, V,\, \theta$ are the net active power injection,  net reactive power injection, voltage magnitude, and voltage angle, respectively. The approximate labelling function $\mathcal{\tilde{F}}$ in eq. \ref{eq:classifier} is trained by minimizing the distance $\|Y - \tilde{Y}\|_{p}$, where $p$ is the distance norm.

\subsection{Operational challenges}
The previous section describes how the training database is generated via TDS and how an ML classifier is trained to predict security labels. This workflow has two issues:
\begin{itemize}
    \item Per scenario model proliferation: In the conventional DSA workflow, a separate classifier has to be trained for each contingency in the contingency list \cite{ren_transfer_2020}. As the number of credible contingencies to consider increases, the number of models to train increases, with the attendant cost of hyperparameter tuning. Each model must also be retrained/updated when new OC data arrives.
    \item Inadequate generalization to unseen scenarios: When there is a new contingency, the current catalogue of classifiers cannot be used. First, labelled training data must be generated for the new contingency through TDS, and then a new classifier can be trained. A change in the original system topology also renders previously trained classifiers invalid.
\end{itemize}
In practice, these two operational challenges are related and become increasingly costly when the list of credible contingencies is large, and there are limited resources for TDS even in the offline phase. A tabular foundation model (TFM) fulfils the desired quality for a sample-efficient DSA model, while the use of EDC encoding in the TFM enables better generalization to unseen contingencies.

\subsection{Generalization to unseen faults}
Let $\mathcal{L}$ be the true labelling function:
\begin{equation}
\label{eq:true-label}
\mathcal{L}(X, \Phi) \rightarrow Y,
\end{equation}
where $\Phi$ is a scenario vector that captures the network topology (admittance matrix) and the fault scenario (fault duration, location, post-fault topology, etc.). $\mathcal{L}$ is a deterministic function of the pre-fault state $X$ and the scenario $\Phi$.
The ML classifier $\mathcal{\tilde{F}}$ learns a conditional approximation $\mathcal{\tilde{F}}(X | \Phi) \simeq \mathcal{L}(X, \Phi)$, where the scenario vector $\Phi$ is implicitly approximated by the classifier. If $\Phi$ is fixed over a distribution $\mathcal{D}$ of $X$, $X$ is a sufficient statistic for approximating $\mathcal{L}$ given sufficient training samples. However, when the scenario changes such that $\Phi_1$ $\neq$ $\Phi_0$, the prediction error: $|\mathcal{L}(X, \Phi_1) - \mathcal{\tilde{F}}_{\Phi_0}(X)|$ is unbounded. For example, $\Phi_1$ could be a new fault location.  Consequently, $\mathcal{\tilde{F}}$ is a poor approximation of $\mathcal{L}$ in this scenario, because the model is extrapolating \cite{arowolo_exploring_2025}. 
A change $\Phi$ in this case is known as concept shift \cite{lu_learning_2018}. Concept shift changes the true labelling function $\mathcal{L}$, which results in a change in the true decision boundary. Importantly, the true decision boundary is theoretically not identifiable without additional $\Phi$ samples or further causal assumptions \cite{ben-david_theory_2010, ben-david_impossibility_nodate}. If $\Phi$ is made part of the observation through a continuous encoding, contingencies which were indistinguishable under concept shift become separable points in the new augmented input space. Generalizing to an unseen contingency becomes interpolation between points rather than extrapolation. In Section IV-E, we investigate how to represent $\Phi$ for good generalization performance.


\section{Tabular Foundation Model for pre-fault DSA}
\begin{figure}[t]
    \centering
    \includegraphics[width=0.48\textwidth]{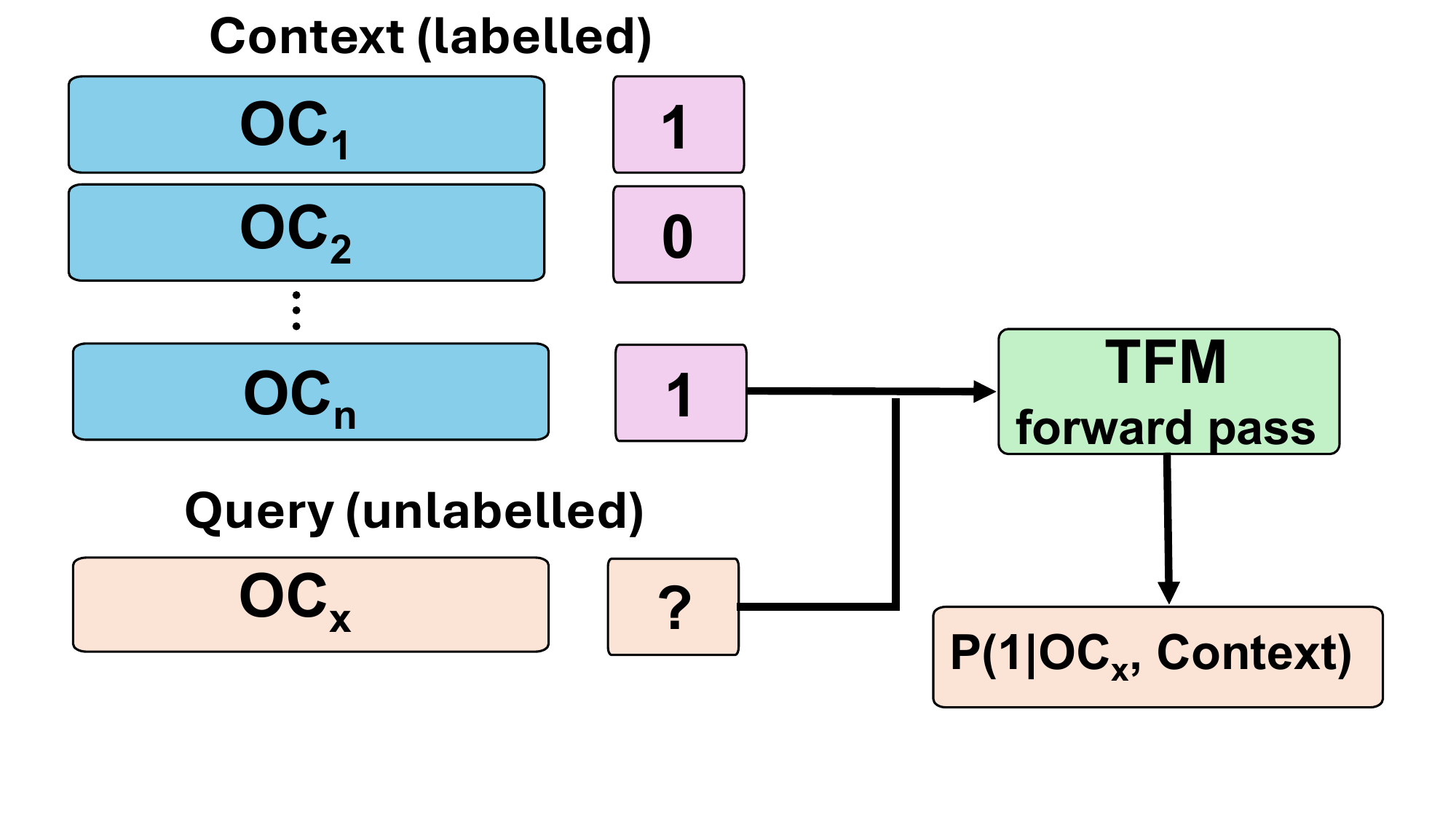}
    \vspace{-4mm}
    \caption{TFM prediction requires no parameter update or hyperparameter tuning}
    \label{fig:TFM_DSA_workflow}
\end{figure}
\vspace{-0.1cm}
\subsection{Background: Tabular Foundation Model}
 We adopt TabPFN \cite{hollmann_accurate_2025} as the TFM in this work. Conventional ML classifiers optimize their parameters for the training data through gradient descent. Whereas the TFM is a transformer pretrained once on 100 million synthetic datasets and can be applied to a new task or dataset without further retraining. During inference, the TFM performs in-context learning (ICL). ICL is a learning framework whereby transformer models perform new tasks simply by processing new input-output pairs without requiring model retraining or any parameter updates \cite{brown_language_2020}. Given the labelled training data $\mathcal{D}_{tr}$ = $\{ (X_i,Y_i)\}^{n}_{i=1}$, and the unlabelled queries, the TFM predicts the labels for all queries in a single forward pass as shown in Fig. \ref{fig:TFM_DSA_workflow}. This is analogous to how large language models (LLMs) infer responses using the provided prompt as context. The training step in conventional classifiers reduces to caching the training data in memory, but no further parameter update or hyperparameter search is required. The TFM approximates the Bayesian posterior predictive distribution: \[ P(Y_{test} | X_{test}, X_{train}, Y_{train})\]. The prediction step is a Bayesian inference conditioned on the available training data.
 The TFM can achieve accurate predictions with a small $n$ and requires only one pretrained model across different tasks. The proposed workflow exploits these two properties.

 \begin{figure}[t]
    \centering
\includegraphics[width=0.485\textwidth]{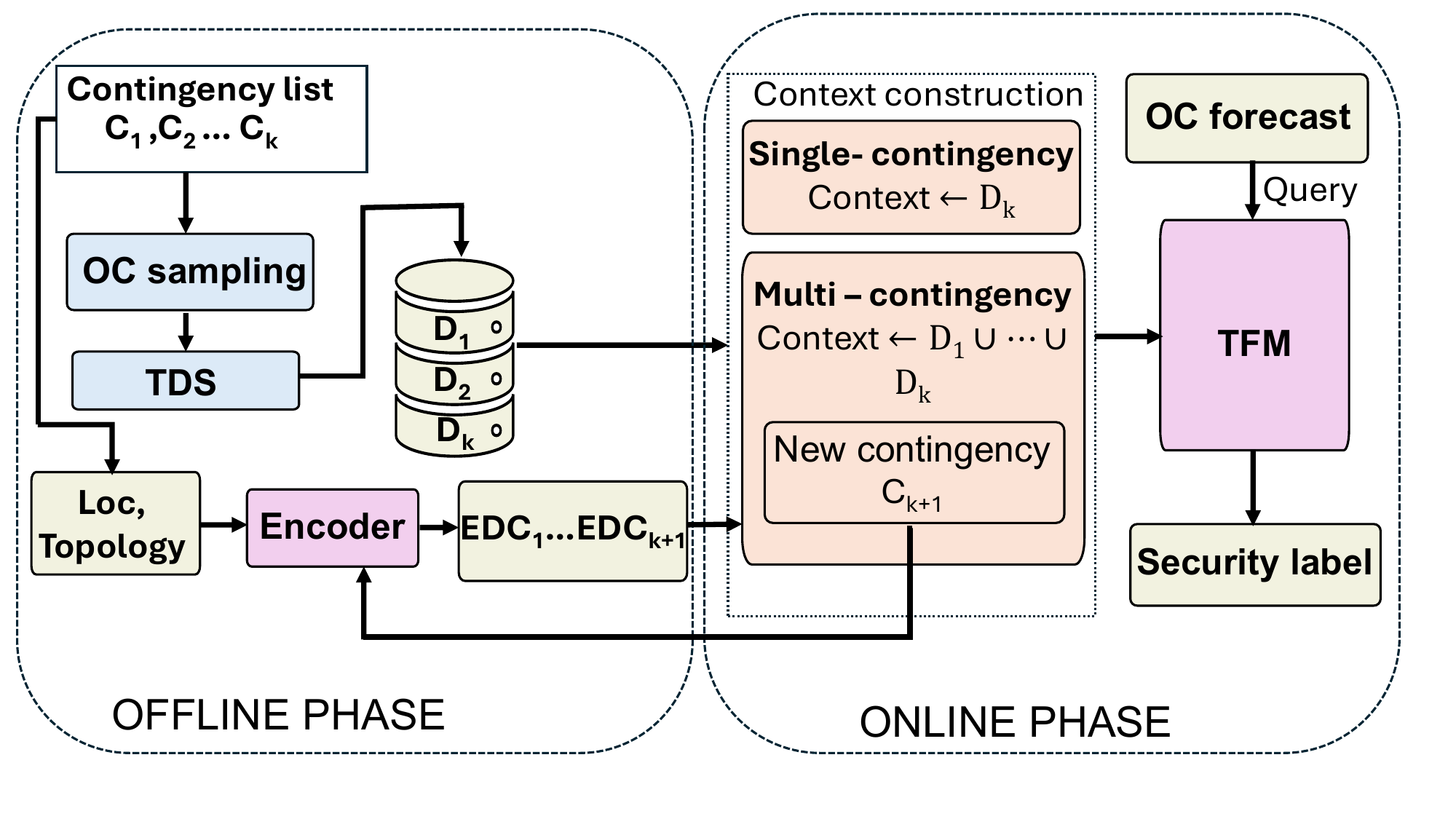}
    \caption{Workflow for the proposed approach. In the offline phase, data is generated via TDS. In the online phase, the TFM may be used to predict labels for individual or multiple contingencies. For unseen contingencies, EDC-based encoding may be provided as additional context for the TFM.}
    \label{fig:complete_workflow}
\end{figure}

\subsection{Proposed workflow}
This section describes the proposed workflow for online pre-fault DSA using TFM as illustrated in Fig. \ref{fig:complete_workflow}. Our proposed approach is comprehensive and addresses challenges related to large database generation and generalization to unseen contingencies. The first step in the workflow is to identify a list of critical contingencies to assess in the online phase. For the selected contingencies, labelled training data is generated via TDS. A wide range of OCs are simulated to ensure maximum diversity in the dataset. In this work, we sample the OCs for training samples using Monte Carlo, but other approaches which produce high-quality TDS-generated labels may be used as well. The labelled training database for each contingency serves as context for the TFM during inference. 
For a large list of contingencies, the use of one TFM simplifies the process of training individual classifiers and reduces it to a one-step prediction via ICL. Depending on available memory resources, TFM allows for inference sequentially or in parallel. For a short list of critical contingencies, a multi-contingency DSA approach may be taken, where training data for each contingency can be part of a pooled context, and a prediction can be made for any contingency in the database. For a new/unseen contingency, the sample efficiency of the TFM can be exploited. We propose two options for the system operator under a preventive DSA framework: 1) a few labelled samples of the unseen contingency can be generated via TDS and added to the TFM context. 2) Where new data generation is impractical, the TFM may be used for zero-shot prediction of the new contingency. Here, we opt for the first option and propose the use of the EDC-based encoding as additional input for each contingency so the TFM may learn structural relationships between known contingencies and the unseen ones.

\subsection{Electrical Distance Coordinate Encoding}
To enhance the generalization of a trained DSA classifier to unseen contingencies, we build a scenario vector that captures the contingency characteristics, such as location or post-fault topology. It is desired that the built representation is continuous and embeds similar scenarios closer in the representational space than dissimilar ones. We rely on the electrical distance coordinate (EDC) proposed by \cite{qi_continuous_2025} to replace the discrete contingency index with a continuous contingency location in a metric space. For a pair of buses $(i, j)$, the electrical distance between them is given as:
\begin{equation}
\label{eq:electrical_distance}
d_{ij} = |z_{ii} + z_{jj} - z_{ij} - z_{ji}|
\end{equation}
where $z_{ii}$ and $z_{jj}$ are self-impedances, while $z_{ij}$ and $z_{ji}$ are mutual impedances. To ensure that Euclidean distances in the coordinate space align with electrical distances, a set of reference nodes, $\mathcal{k}$, is selected that minimizes the distance residual:
\begin{equation}
\label{eq:residual_metric}
R = \sum_{i < j} (|| c_{i} - c_{j}||_{2} - d_{ij})^{2}
\end{equation}
where $c_{i} = [d_{i,r1} ... d_{i,rk}]$ is the coordinate of bus $i$, found using electrical distance from $i$ to each of the reference nodes. We used 8 reference nodes in our implementation. We select reference nodes using greedy forward selection. Each contingency's scenario vector is the pre-fault electrical distance coordinate of the faulted bus; $EDC_{n} \in \mathds{R}^{8}$ for $n \in \{1 ... 22\}$. We make two observations. First, the EDC for a new/unseen contingency is continuous relative to the training contingencies. Second, only the contingency location and Z-bus matrix of the original network are needed to compute EDC. However, two contingencies at the same bus which are cleared by tripping different lines would have identical EDC encoding. This limits the ability of a classifier using EDC encoding to distinguish these two faults.
To this end, we propose the mod-EDC encoding, which takes the post-fault topology into account. We assume the clearing line for a contingency is known, and the clearing times are the same for all contingencies. Our approach is motivated by the classic equal area criterion (EAC) analysis for transient stability. In EAC analysis, the system's response is determined by the accelerating area and the decelerating area. The accelerating area largely depends on fault location and clearing time, while the decelerating area depends on the synchronizing power of the network after the fault is cleared. Therefore, we use the post-fault topology to compute the differential coordinate:
\begin{equation}
\label{eq:differential_metric}
Q_{n} = EDC_{n}^{pre} - EDC_{n}^{post}
\end{equation}
$Q_{n}$ is a label-free proxy measuring how severely the clearing action changes the synchronizing power of the network. The modified EDC encoding concatenates eight pre-fault EDC values with eight pre-/post-fault differences and is given by:
\begin{equation}
\label{eq:mod-EDC}
M_{n} = [EDC_{n}^{pre} || Q_{n}] \in \mathds{R}^{16}
\end{equation}
For EDC and mod-EDC encodings, the 8 reference nodes are at buses: 15, 5, 49, 48, 1, 53, 47 and 26 of the test system. The number of reference nodes is fixed and is not optimized for the test system.

\subsection{Baselines}
We compare the proposed approach to baselines from the literature. The baselines consist of classical ML models and deep learning models. For deep learning baselines, we have considered FT-Transformer \cite{gorishniy_revisiting_2023} and GATv2 \cite{brody_how_2022}. Similar to the TFM, FT-Transformer uses a tabular transformer but is trained from scratch on the training data, without any pretraining. The GATv2 implicitly encodes the system graph topology through learned attention weights. The baseline models and the hyperparameter settings are shown in Tab.  \ref{tab:hparam_grid}. For each contingency baseline, 10 configurations are sampled randomly from the hyperparameter search space with 3-fold cross-validation for each configuration. This gives an equal budget of 30 hyperparameter tuning rounds per baseline model. For discrete-valued hyperparameters, we searched randomly over the grid search space. For continuous-valued hyperparameters, we used log-uniform sampling over the search space.
\begin{table}[t]
  \centering
  \setlength{\tabcolsep}{4pt}
  \footnotesize
  \caption{Hyperparameter search spaces for the baseline models.} 
  \label{tab:hparam_grid}
  \begin{tabularx}{\columnwidth}{@{}l l >{\raggedright\arraybackslash}X l@{}}
    \toprule
    Model & Hyperparameter & Search space & Sampling \\
    \midrule
    \multirow{4}{*}{XGBoost}
      & \texttt{n\_estimators}  & \{100, 200, 400\}    & discrete \\
      & \texttt{max\_depth}     & \{3, 4, 6, 8\}       & discrete \\
      & \texttt{learning\_rate} & \{0.03, 0.1, 0.3\}   & discrete \\
      & \texttt{subsample}      & \{0.7, 1.0\}         & discrete \\
    \addlinespace
    \multirow{3}{*}{MLP}
      & \texttt{hidden\_layer\_sizes}
        & \{(64), (128), (128,~64), (256,~128)\} & discrete \\
      & \texttt{alpha}                & $[10^{-5}, 10^{-2}]$ & log-uniform \\
      & \texttt{learning\_rate\_init} & $[10^{-4}, 10^{-2}]$ & log-uniform \\
    \addlinespace
    \multirow{3}{*}{DT}
      & \texttt{max\_depth}         & \{3, 5, 8, 12, 20\} & discrete \\
      & \texttt{min\_samples\_leaf} & \{1, 2, 5, 10\}     & discrete \\
      & \texttt{criterion}          & \{gini, entropy\}   & discrete \\
    \addlinespace
    \multirow{2}{*}{SVM-RBF}
      & \texttt{C}     & $[10^{-2}, 10^{3}]$ & log-uniform \\
      & \texttt{gamma} & $[10^{-4}, 10^{0}]$ & log-uniform \\
    \addlinespace
    \multirow{7}{*}{FT-Transformer}
      & \texttt{d\_token}      & \{32, 64, 128\}           & discrete \\
      & \texttt{n\_blocks}     & \{1, 2, 3\}               & discrete \\
      & \texttt{n\_heads}      & \{4, 8\}                  & discrete \\
      & \texttt{attn\_dropout} & \{0, 0.1, 0.2\}           & discrete \\
      & \texttt{ffn\_dropout}  & \{0, 0.1, 0.2\}           & discrete \\
      & \texttt{lr}            & $[10^{-4}, 10^{-3}]$      & log-uniform \\
      & \texttt{weight\_decay} & $[10^{-6}, 10^{-3}]$      & log-uniform \\
    \addlinespace
    \multirow{6}{*}{GATv2}
      & \texttt{hidden}        & \{16, 32, 64\}            & discrete \\
      & \texttt{heads}         & \{2, 4\}                  & discrete \\
      & \texttt{n\_layers}     & \{2, 3\}                  & discrete \\
      & \texttt{dropout}       & \{0, 0.1, 0.2\}           & discrete \\
      & \texttt{lr}            & $[3\times10^{-4}, 3\times10^{-3}]$ & log-uniform \\
      & \texttt{weight\_decay} & $[10^{-6}, 10^{-3}]$      & log-uniform \\
    \addlinespace
    MDSA-S & $\sigma_3,\sigma_4,\sigma_5$
      & $\{10^{-5}, \dots, 10^{5}\}$  & axis-sweep \\
    \bottomrule
  \end{tabularx}
\end{table}
\section{Case studies}
Based on the proposed workflow, we design case studies to answer 3 questions:
\begin{enumerate}
    \item How sample efficient is the proposed approach compared to existing benchmarks using the standard one-classifier-per-contingency workflow?
    \item How well does the proposed approach perform when a single model is used for assessing multiple contingencies?
    \item When, and how well, does a continuous EDC-based encoding enable the TFM to generalize to unseen contingencies, and how much additional labelled data from the unseen contingency is needed for reliable generalization?
\end{enumerate}
Positive answers to the above questions would position the proposed approach as a viable solution to the two identified challenges.
After describing the test system and modelling assumptions, we evaluate the proposed approach by comparing its performance with six baselines at different training database sizes. Subsequently, we evaluate the robustness of the proposed approach to severe class imbalance and compare it with a state-of-the-art multi-label classification method. We continue by exploring the generalization of the proposed approach to unseen contingencies and finish by considering computational aspects of the proposed approach.

\begin{figure}[t]
    \centering
    \includegraphics[width=0.48\textwidth]{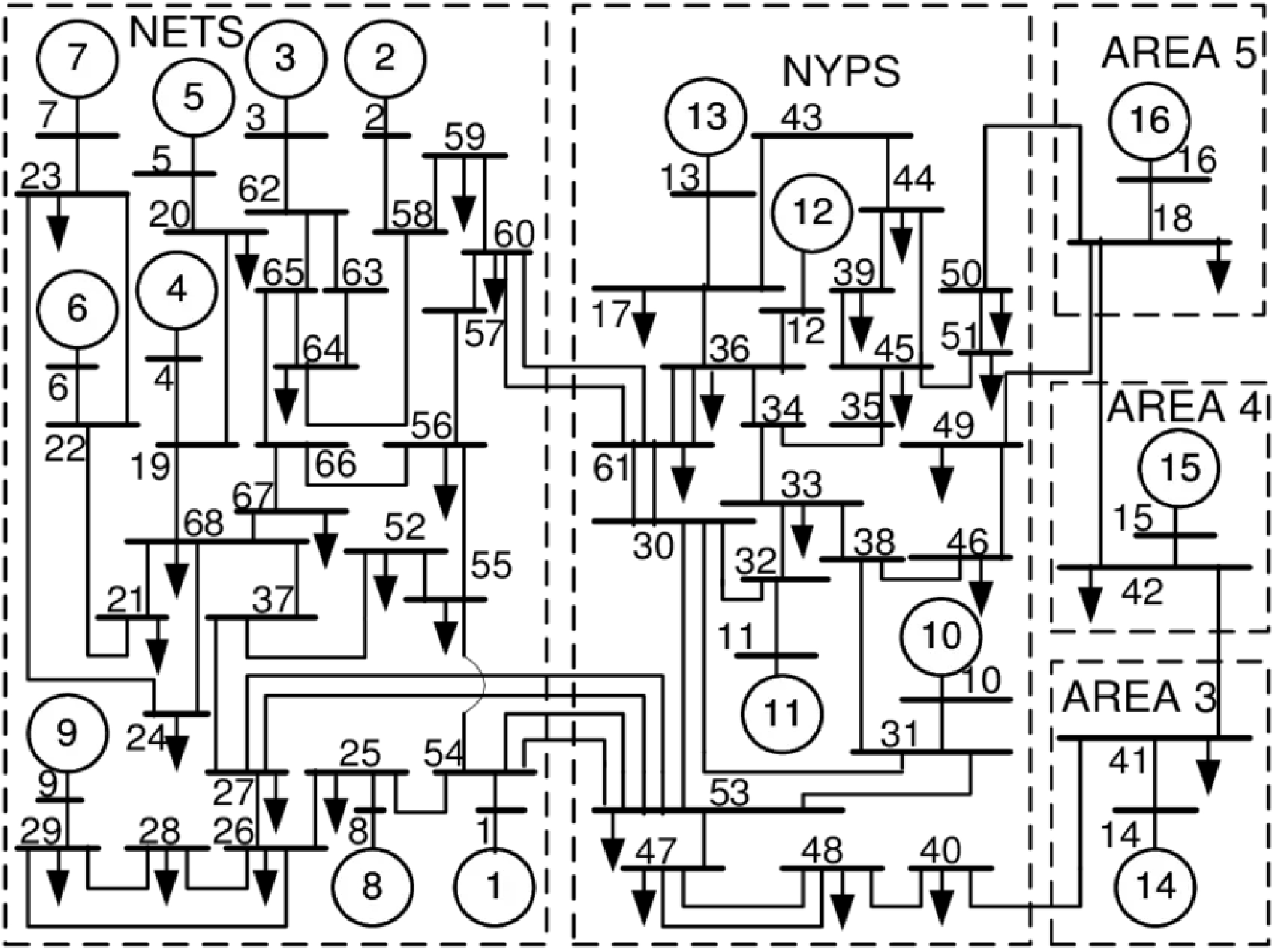}
    \vspace{-1mm}
    \caption{The test system is the IEEE 68-bus network}
    \label{fig:test_network}
\end{figure}

\subsection{Test system and  Assumptions}
The IEEE 68-bus system shown in Fig. \ref{fig:test_network} was used as the test system. A set of 120,000 samples was generated for each of the 22 contingencies considered. The samples here describe the system under AC steady-state operation. The transient stability of the 22 contingencies was simulated in the time domain for 10s. Using the average of 0.2s per simulation reported in \cite{cremer_optimization-based_2019} for the same test system and set of contingencies, generating the 120,000 samples for each contingency takes about 6.7 CPU hours and about 147 CPU hours for the whole dataset of 22 contingencies. Each contingency corresponds to a 3-phase bus fault, and the fault is cleared after 0.1s by tripping a line, leading to an N-1 post-contingency network topology. The stability label is 1 if TSI \textgreater 0 from Eq. \ref{eq:tsi}, i.e., the maximum power angle difference between any two generators remains below 360$^\circ$ over the 10s simulation, indicating the system remains stable. Otherwise, the system is unstable, and Y is 0. The simulations are performed in MATLAB R2016b Simulink. We use the IEEE-68 bus dynamic model as specified in \cite[App. A-B]{bikash_robust_2005}, including the generator, exciter, power system stabilizer, and load models. In the TDS, all loads are modelled as constant impedances, as specified in \cite{bikash_robust_2005}. The list of the selected contingencies is given in Tab. \ref{tab:fault_data}. 
To create high variability in the OCs, we have used the same sampling procedure as in \cite{cremer_optimization-based_2019}. The active loads are sampled from a multivariate Gaussian distribution with a Pearson’s correlation coefficient of
0.75 between all load pairs. Loads were converted to a marginal Kumaraswamy(1.6, 2.8) distribution by making use of the inverse transformation method, then
scaled to be within $\pm 50\%$ of the nominal active loads. The reactive loads were scaled from their nominal values by assuming constant bus impedances. The total active power was uniformly distributed across all generators, then normalized with respect to their capacity-sharing ratio, with the reactive powers distributed in the same way. To ensure AC feasibility, each sampled loading condition was projected onto the AC-feasible set by solving a nonlinear AC optimal power flow problem. The objective minimizes the sum of absolute deviations of each generator's active and reactive power from its sampled share, subject to the AC power balance equations, generator power limits, and generator power factor within [0.75, 1], while loads remain fixed. The problem was solved using IPOPT 3.12.4 \cite{cremer_optimization-based_2019}. The solution of the optimization problem is the initial steady-state condition for the TDS. The data generation process produces 22 datasets where the ratio of stable to unstable observations ranges from 0.09 to 0.99, with a mean of 0.56. All simulations produce a stability label, which we assume to be the ground truth label.

\begin{table}[t]
\centering
\caption{Selected contingencies and label distribution}
\label{tab:fault_data}
\setlength{\tabcolsep}{4pt}
\footnotesize
\begin{tabular}{c c c c c}
\hline
Fault & Faulted & Clearing line & Clearing line & Stable \\
ID    & Bus     & From   & To     & Proportion \\
\hline
\hline
1  & 23 & 23 & 22 & 0.90 \\
2  & 23 & 24 & 23 & 0.89 \\
3  & 22 & 23 & 22 & 0.85 \\
4  & 62 & 63 & 62 & 0.96 \\
5  & 62 & 65 & 62 & 0.99 \\
6  & 25 & 25 & 54 & 0.30 \\
7  & 25 & 26 & 25 & 0.30 \\
8  & 58 & 58 & 57 & 0.93 \\
9  & 58 & 59 & 58 & 0.93 \\
10 & 29 & 29 & 28 & 0.09 \\
11 & 29 & 29 & 26 & 0.10 \\
12 & 54 & 55 & 54 & 0.90 \\
13 & 54 & 54 & 53 & 0.52 \\
14 & 54 & 25 & 54 & 0.50 \\
15 & 17 & 43 & 17 & 0.27 \\
16 & 36 & 36 & 61 & 0.29 \\
17 & 36 & 36 & 34 & 0.14 \\
18 & 32 & 32 & 30 & 0.62 \\
19 & 32 & 33 & 32 & 0.36 \\
20 & 31 & 31 & 30 & 0.54 \\
21 & 31 & 31 & 53 & 0.56 \\
22 & 31 & 38 & 31 & 0.39 \\
\hline
\end{tabular}
\end{table}

Each OC is represented by four steady-state quantities per bus: active power injection, reactive power injection, voltage magnitude and phase angles, resulting in a 272-dimensional feature vector. We run each experimental workflow and training size combination under four random seeds and report the mean $\pm$ std of the selected metrics. We sample training subsets from the original pool of 120,000 OCs. We use stratified sampling on the binary target to preserve the target class proportions in the training datasets. We set a minimum of 1 sample per target class so that each class is represented in the training data at every data size. This prevents degenerate cases with data from only one class for datasets with severe class imbalance. We uniformly sample a separate test set of 12,000 samples. We have intentionally chosen a large test data size relative to the training data sizes to reliably evaluate the model's generalization to new OCs within the same distribution. We ensure the same test set is used across different training dataset sizes for each seed. This ensures the evaluation is fair with identical test distributions at every size.

We have used \textit{TabPFNv2.6} as the TFM in this work. We implemented the baseline classifiers with \textit{scikit-learn}, \textit{PyTorch}, and \textit{XGBoost}. All experiments were performed on an AMD EPYC-7H12 computer with 64 CPU cores. Deep learning models, including the TFM, were run on an Nvidia H100 GPU with 94GB memory and CUDA 12.1.1.


\subsection{Evaluation metrics}
In this section, we introduce the four metrics used to evaluate the various models. Given that the DSA prediction task is a binary classification problem, we can obtain four types of results for each prediction: (1) A true positive (TP - secure case correctly predicted), (2) A false positive (FP - insecure case wrongly predicted as secure), (3) A true negative (TN - insecure case correctly predicted), (4) A false negative (FN - secure case wrongly predicted as insecure). The evaluation metrics are:
\begin{itemize}
    \item Balanced accuracy: The average recall across classes, which corrects for class imbalance. Unlike simple accuracy, it does not get artificially inflated by the majority class and is better suited to tasks such as DSA, where insecure cases are rare:
    \begin{equation}
    \text{Balanced Accuracy}
    = \frac{1}{2}\left(\frac{TP}{TP + FN} + \frac{TN}{TN + FP}\right)
    \end{equation}
    \item Precision: This is the proportion of samples predicted as positive (secure) which are actually positive. A high precision means secure predictions are more reliable:
    \begin{equation}
    \text{Precision} = \frac{TP}{TP + FP}
    \end{equation}
    \item Specificity: This is the proportion of actually negative (insecure) samples that are correctly identified as negative. A high specificity indicates the model misses fewer insecure cases. A high specificity is the most operationally critical metric for DSA:
    \begin{equation}
    \text{Specificity} = \frac{TN}{TN + FP}
    \end{equation}
    \item Macro F1: The F1 score is the harmonic mean of precision and recall. Macro F1 computes the F1 score for each class separately and then takes the average. This ensures each class contributes equally regardless of its frequency:
    \begin{equation}
    F1_c
    = \frac{2\,TP_c}{2\,TP_c + FP_c + FN_c}
    \end{equation}
    Macro average over $C=2$ classes:
    \begin{equation}
    \text{Macro F1} = \frac{1}{2}\sum_{c=1}^{2} F1_c
    \end{equation}
\end{itemize}

\begin{figure*}[t]
    \centering
    \includegraphics[width=\textwidth]{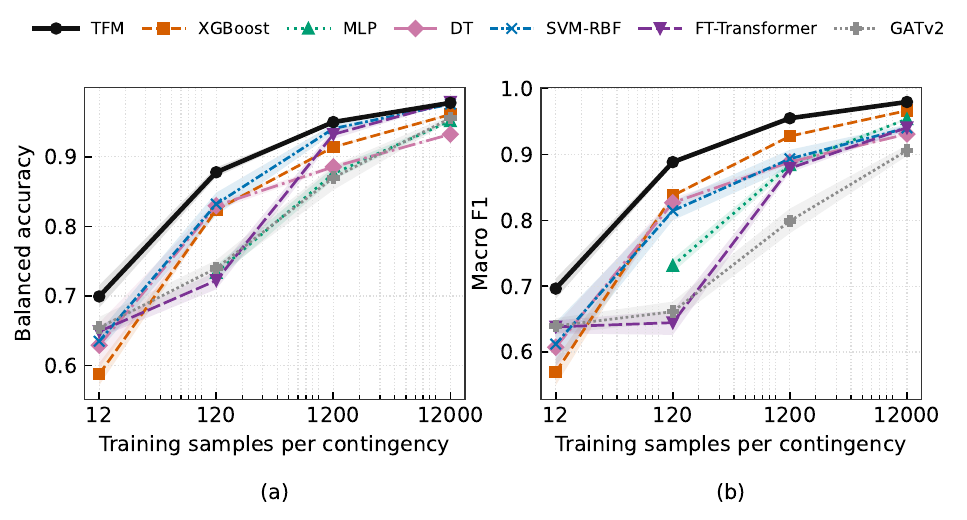}
    \vspace{-5mm}
    \caption{Average Balanced accuracy (a) and Macro F1 (b) of the TFM compared to the baseline models. 1 standard deviation is shown in the shaded band. The TFM has better balanced accuracy than all six baselines at the three lower training dataset sizes and similar balanced accuracy as FT-Transformer with 12000 training samples. TFM also has better Macro F1 than all baselines across all the tested training dataset sizes.}
    \label{fig:balanced_accuracy_macro_f1}
    \vspace{-0.1cm}
\end{figure*}

\subsection{Predictive performance of per-contingency workflow} 
To test the sample efficiency and predictive performance of the TFM, we train a separate classifier for each of the 22 contingencies. We consider training budgets $n$ of orders of magnitudes, $n \in \{12, 120, 1200, 12000\}$. A training size of 12 represents a scarce data scenario where only a few TDS can be performed. On the other hand, a training size of 12000 is consistent with typical training sizes in the literature \cite{zhang_confidence-aware_2021,cremer_optimization-based_2019,ren_pre-fault_2023}. The six baseline classifiers are tuned using 10 hyperparameter combinations evaluated with 3-fold cross-validation. This results in 30 model variants per combination of model, contingency, training size and random seed. The best model is selected based on its balanced accuracy and trained with the full training data split. Since the TFM requires no hyperparameter tuning, we simply fit one classifier per contingency, training size and random seed combination.

Fig. \ref{fig:balanced_accuracy_macro_f1}a shows the average balanced accuracy across training dataset sizes. TFM shows significantly higher balanced accuracy at smaller TDS data budgets, benefiting from the strong prior the model provides. At the biggest data budget, the gap to the baseline models gets smaller. TFM (97.7\%) remains comparable to FT-Transformer (97.8\%), the best baseline at 12000 training samples. We note that an MLP classifier could not be trained at the smallest budget of just 12 samples, as it led to severe overfitting and a degenerate model. The advantage of TFM over the baselines is even more pronounced for Macro F1. Macro F1 balances the operational need for high precision (confidence in secure prediction) and high specificity (low missed alarms). The TFM achieves an average Macro F1 score of about 90\% with just 120 training samples, as shown in Fig. \ref{fig:balanced_accuracy_macro_f1}b. It outperforms the closest baseline \textit{XGBoost} by about 5\% on average. Other baselines, apart from \textit{XGBoost}, require an order of magnitude more labelled data to reach a similar performance.

\begin{figure}[t]
    \includegraphics[width=0.48\textwidth]{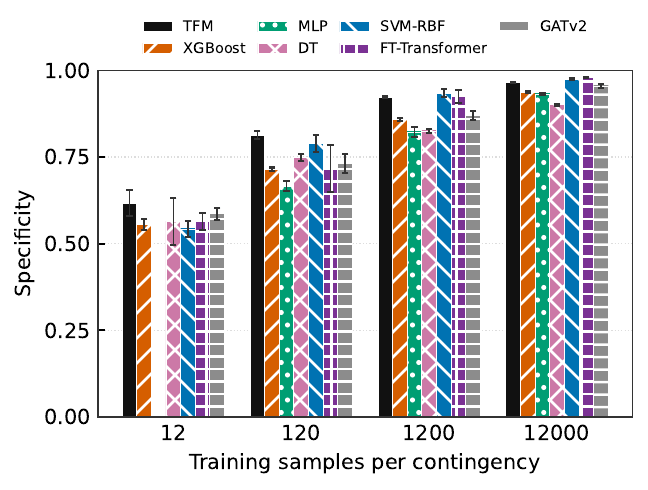}
    \vspace{-1mm}
    \caption{\footnotesize Average specificity of the TFM and the baselines at different training budgets. Error bars show 1 standard deviation over the four random seeds. The TFM's specificity is the highest at 12 and 120 training samples. SVM-RBF and FT-Transformer have slightly higher specificity at 1200 and 12000 training samples.}
    \label{fig:specificity_bars}
\end{figure}

Since correctly identifying insecure cases is more operationally important than correctly identifying secure ones, we focus on the specificity of the classifiers. Fig. \ref{fig:specificity_bars} shows specificity (insecure-class recall) across the 4 data budgets. TFM has the highest specificity with only 12 training samples, about 3\% higher than GATv2, the best baseline at that training data size. With 120 samples, TFM reaches an average specificity of 81\%, with the closest baseline being SVM-RBF at 79\%. SVM-RBF (97.5\%) and FT-Transformer (97.9\%) slightly outperform TFM (96.4\%) with 12000 samples, but they trade off higher specificity with lower secure class recall (sensitivity). This is because SVM-RBF operates with a margin-based objective and FT-Transformer is trained with a class-balanced loss. Both objectives can be adjusted to prioritize specificity over sensitivity. TFM is the only model which correctly classifies more than 80\% of all insecure cases on average when only 120 training samples are available.

\begin{figure}[t]
    \includegraphics[width=0.48\textwidth]{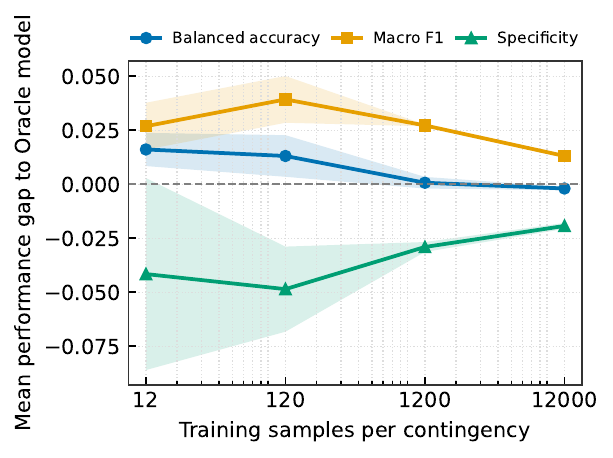}
    \vspace{-1mm}
    \caption{\footnotesize Mean performance gap between the TFM and an oracle model which chooses the best of the six baseline models for each contingency. The positive axis indicates the TFM is better, and the negative axis indicates the oracle is better.}
    \label{fig:gap_to_oracle}
\end{figure}

Finally, we compare TFM to an oracle model that picks the best-performing baseline classifier for each contingency. Fig. \ref{fig:gap_to_oracle} shows the mean gap between the oracle model and the TFM across all training data budgets for balanced accuracy, Macro F1 and Specificity. The TFM outperforms the oracle model on balanced accuracy and Macro F1 but is inferior on Specificity. We note that the oracle model is an idealistic model that cannot be deployed in practice since it picks the best-performing model on the test dataset for each contingency. This is different from a classic ensemble model. The TFM still matches or outperforms the oracle model using the best available baseline, with zero fine-tuning. The predictive advantage of TFM over the baselines is more pronounced at small data budgets and 'harder' contingencies with high class imbalance, demonstrating its advantage in realistic operational scenarios.
\subsection{multi-contingency DSA classifier}
This case study tests whether a single model can assess \textit{k} known contingencies simultaneously. We compare the use of a multi-label learning-based DSA approach (MDSA-S) \cite{ren_pre-fault_2023} to a single pooled TFM applied to all 22 contingencies simultaneously. For both approaches, we assume the training data is fully labelled. We use the same set of 1200 OCs as training samples for each contingency. For the TFM approach, we added a one-hot encoding of fault ID to the feature vector to distinguish between faults. This brings the feature dimension to 294. 
\begin{figure*}[t]
    \centering
    \includegraphics[width=\textwidth]{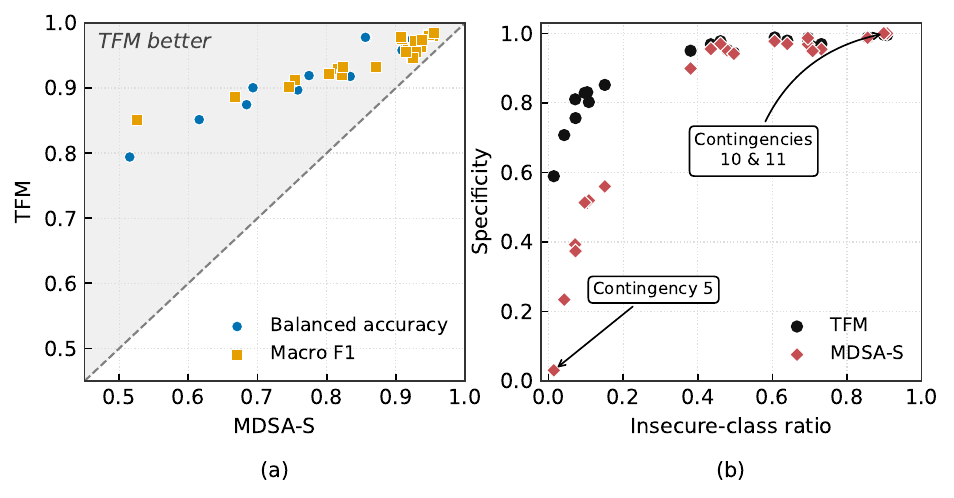}
    \vspace{-5mm}
    \caption{Average performance comparison of TFM and the MDSA-S baseline. (a) Pairwise scatterplot of average balanced accuracy and Macro F1 for TFM versus MDSA-S. The diagonal shows where both methods perform identically. TFM consistently outperforms MDSA-S for all 22 contingencies. (b) Specificity versus insecure-class ratio for all 22 contingencies. TFM shows robustness for contingencies where insecure cases are rare.}
    \label{fig:hypothesis2}
    \vspace{-0.1cm}
\end{figure*}
We fit the TFM on a combined \textit{1200 x 22} context rows. The MDSA-S approach jointly couples the 1200 OC inputs with the complete 22-column label matrix. 
We train the model using the same proximal accelerated gradient approach as in \cite{ren_pre-fault_2023}.
We test both approaches using a separate test set of 12000 OCs per contingency. We report mean metrics across 4 seeds, averaged across all contingencies.
Fig. \ref{fig:hypothesis2}a compares the performance of the pooled TFM and the MDSA-S approach. Across all 22 contingencies, the pooled TFM achieves higher balanced accuracy and Macro F1 than the MDSA-S approach. Fig. \ref{fig:hypothesis2}b shows the specificity of both approaches for all the contingencies. In particular, we focus on 'harder' contingencies where the insecure class ratio is small. Again, the pooled TFM is robust, with worst-case specificity of 58.9\% and worst-case Macro F1 score of 85.1\% for contingency 5, where only 1\% of samples are insecure, as shown in Fig. \ref{fig:hypothesis2}b. In contrast, MDSA-S's specificity degrades to 3.1\% with a Macro F1 score of 52.6\%. This indicates that for contingencies with rare insecure cases, it defaults to predicting secure cases and misses relevant insecure cases. However, MDSA-S closes the gap to TFM on specificity when the training data has a balanced proportion of secure and insecure cases. We elaborate on this behaviour from the MDSA-S objective function \cite{ren_pre-fault_2023}.
The objective of MDSA-S is to minimize:
\begin{align}
\min_{W_x, W_y} \; \frac{1}{2} \left\| X W_x + Y W_y - Y \right\|_F^2 
+ \sigma_3 \left\| W_x \right\|_1  \nonumber\\
+ \sigma_4 \left\| W_y \right\|_1
+ \frac{\sigma_5}{2} \operatorname{tr}\!\left( (1 - R) W_x^\top W_x \right)
\end{align}
where $\sigma_3, \sigma_4$, and $\sigma_5$ are hyperparameters of the optimization. $W_x$ and $W_y$ are model parameters of the features and labels, respectively. $R$ is the label correlation matrix between different contingencies. $X$ and $Y$ are features and labels, respectively. The collapse in specificity for contingencies with severe class imbalance follows from minimizing:
\begin{equation}
\frac{1}{2} \left\| X W_x + Y W_y - Y \right\|_F^2.
\end{equation}
For binary labels $Y$, this becomes a least-squares estimator of the conditional expectation $\mathbb{E}[y \mid x] = P(y = 1 \mid x)$. For a contingency with severe class imbalance, where the ratio of insecure classes $k \ll 1$, the loss gradient $X^\top (X w - y)$, is dominated by $(1-k)n$ residuals from the secure class and only a small fraction $kn$ of residuals from the insecure class. Therefore, the optimization leaves the decision boundary where the majority class are correctly predicted, and the model learns to ignore the minority class. Since the objective is symmetric in the two classes, it weights residuals from both classes equally, leading to proper discrimination only when the two classes are fairly balanced. When the majority class is insecure, the model correctly predicts the majority class, leading to high specificity but poor accuracy. This is also shown in Fig. \ref{fig:hypothesis2}b for contingencies 10 and 11, where the insecure class ratios are 91\% and 90\%, respectively. We summarize the mean performance across all contingencies in Tab. \ref{tab:hypothesis_2}.

\begin{table}[b]
\centering
\caption{Comparison of pooled TFM and MDSA-S (mean $\pm$ std)}
\label{tab:hypothesis_2}
\setlength{\tabcolsep}{4pt}
\footnotesize
\begin{tabular}{c c c c}
\hline
\textbf{Metric}    & \textbf{TFM (\%)}     & \textbf{MDSA-S (\%)}   & \textbf{Improvement (\%)}\\
\hline
\hline
Precision          & 96.56 $\pm$ 0.15 & 94.87 $\pm$ 0.29 & +1.69 $\pm$ 0.22 \\
Specificity        & 90.07 $\pm$ 0.58 & 75.66 $\pm$ 1.04 & +14.41 $\pm$ 0.71 \\
Macro F1           & 94.43 $\pm$ 0.19 & 85.99 $\pm$ 0.41 & +8.44 $\pm$ 0.38 \\
Balanced accuracy  & 93.50 $\pm$ 0.34 & 83.43 $\pm$ 0.36 & +10.07 $\pm$ 0.31 \\
\hline
\end{tabular}
\end{table}

\subsection{  Contingency generalization}
\begin{figure}[t]
    \centering
    \includegraphics[width=0.48\textwidth]{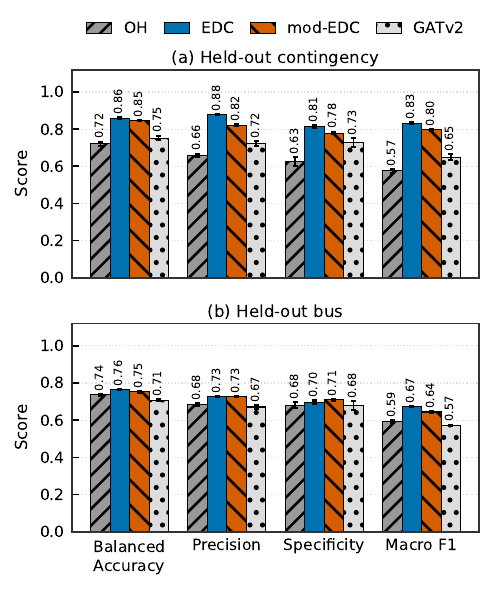}
    \vspace{-1mm}
    \caption{The EDC and mod-EDC encodings outperform both OH encoding and GATv2 for held-out contingencies. EDC and mod-EDC performance degrades when generalizing to contingencies from a held-out bus but remains better than OH encoding and GATv2.}
    \label{fig:hypothesis3_zeroshot}
\end{figure}

\begin{figure}[t]
    \centering
    \includegraphics[width=0.48\textwidth]{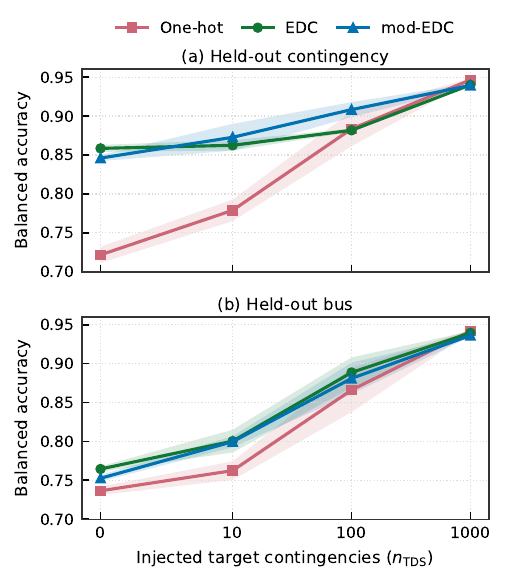}
    \vspace{-1mm}
    \caption{For held-out contingencies, the proposed mod-EDC encoding enables better few-shot generalization with 10 to 100 additional labelled samples. With 1000 additional samples, all 3 approaches converge to the same accuracy. For new contingencies from a held-out bus, EDC and mod-EDC hold a relatively small generalization advantage over OH encoding with few additional labelled samples.}
    \label{fig:hypothesis3_fewshot}
\end{figure}

This case study examines the TFM's generalization to a new, unseen contingency. As described in Section II.C, the system's steady-state features need to be augmented with a scenario vector to distinguish between contingencies. First, we compare the impact of one-hot (OH) encoding, electrical distance coordinate (EDC) encoding and the proposed mod-EDC encoding as the scenario vector when generalizing to an unseen contingency. The above encodings are also compared to the GATv2 model, which directly encodes the graph topology. We consider two generalization scenarios: held-out contingency and held-out bus generalization. For the held-out contingency generalization, we use the multi-contingency DSA approach, whereby we consider 21 contingencies in the labelled training data and one held-out contingency for testing. First, we randomly sample 2000 x 21 training OCs, for an initial total of 42,000 samples. We further downsample to 10,000 of the available 42,000 training samples to reduce memory usage and inference time during prediction. For the held-out bus generalization, we aggregate the 22 contingencies by faulted bus into 11 groups. For each group, all the contingencies at one held-out bus are removed from the model's initial context/training data. This ensures no held-out bus has its EDC/mod-EDC encoding in the model context/training data and generalization to an unseen fault location can be measured. Note that the number of training/context samples in both generalization scenarios is the same (10,000). For all approaches in both generalization scenarios, we predict labels for a separate set of 12,000 OCs from the held-out contingency in a single forward pass of the TFM. 
Fig. \ref{fig:hypothesis3_zeroshot} compares the performance of the three encoding approaches and GATv2 across all four metrics in the held-out contingency and held-out bus generalization scenarios. From the figure, the electrical distance encodings, EDC and mod-EDC, generalize much better to an unseen contingency than OH encoding and GATv2.
However, for a new contingency from an unseen bus/fault location, both encodings drop significantly in performance and remain only marginally better than OH encoding and GATv2. The poor generalization of OH encoding can be attributed to its discrete values. The encoding for an unseen contingency occupies an orthogonal space relative to the training data; hence, the model's accuracy declines because it cannot learn useful connections between the training data and the new scenario. For TFM, which works by ICL, the pre-trained transformer predicts for new rows by 'attending' to context rows which are most similar to the query row. To predict the held-out contingency, the query structure is $(X_{h}, EDC_{h})$. $EDC_{h}$ is present in the context rows if there is another contingency from the same location in the training data. Therefore, TFM uses the decision boundary of the contingency that occurs at electrically similar locations to predict the new contingency. Contingencies which occur at the same location may have similar dynamic behavior \cite{qi_continuous_2025}. When the new fault occurs at a previously unseen location, as in the held-out bus scenario, EDC and mod-EDC still outperform OH and GATv2; however, prediction accuracies are relatively similar in this scenario, highlighting that the generalization of the TFM to completely new fault locations remains limited.
Since the TFM is sample efficient and requires no fine-tuning, we consider a few-shot experiment where a few labels may be generated for the new fault. Here, we assume $n \in \{10,100,1000\}$ labelled data may be generated for the new contingency. We randomly sample $n$ additional OCs for the new contingency to augment the training data. Fig. \ref{fig:hypothesis3_fewshot} shows the average balanced accuracy for the held-out contingency and held-out bus generalization scenarios for OH, EDC and mod-EDC approaches.
When there is no labelled data for the new contingency (zero-shot), EDC is the strongest label-free encoding. However, we observe a crossover point at 10 additional labelled samples, where mod-EDC outperforms EDC. Hence, mod-EDC helps only when a few labelled samples are available for few-shot adaptation. We argue that TFM can quickly adapt its predictions with just a few samples of the new contingency because it can accurately estimate the label distribution $P(y)$ with as few as 10 samples and thus shift the decision boundary appropriately. When the new contingency comes from a previously unseen bus/location, EDC and mod-EDC achieve similar performance.

\begin{figure}[t]
    \centering
    \includegraphics[width=0.48\textwidth]{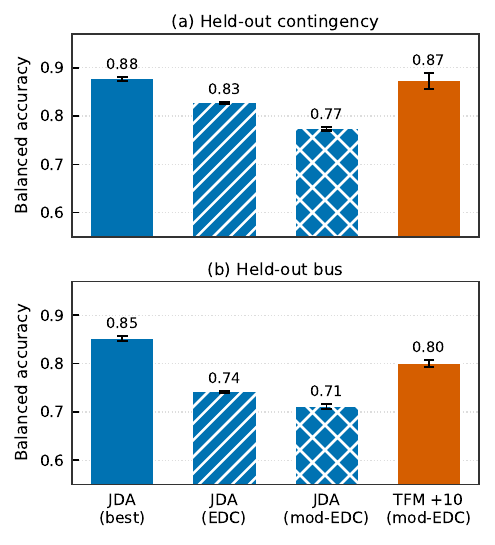}
    \vspace{-1mm}
    \caption{Balanced accuracy of JDA with the three source selection approaches and of TFM with few-shot learning (TFM+10). TFM+10 has a comparable balanced accuracy to the best transfer learning approach via joint domain adaptation (JDA-best) on held-out contingencies. For new contingencies from held-out buses, TFM+10's accuracy is lower than JDA-best, but it still outperforms the two deployable JDA approaches. JDA-best is not deployable in practice, since the best source for each target contingency cannot be determined without target labels.}
    \label{fig:hypothesis3_jda}
\end{figure}

\begin{figure}[t]
    \centering
    \includegraphics[width=0.48\textwidth]{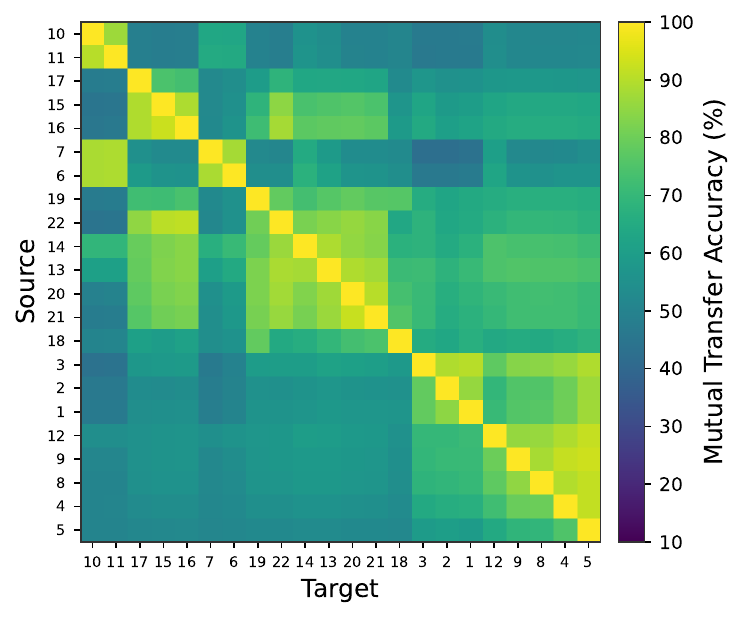}
    \vspace{-1mm}
    \caption{Mutual transfer accuracy (MTA) between all contingency pairs.}
    \label{fig:heatmap_jda}
\end{figure}

Finally, we compare the generalization performance of TFM with that of the transfer learning approach in \cite{ren_transfer_2020}. The transfer learning approach employs joint domain adaptation (JDA) between 2 contingency pairs, assuming the new/unseen contingency is related to a known one in the training database. The known contingency is the source domain, while the new contingency is the target. We have used XGBoost, with fixed hyperparameters from the first study, as the DSA model in our JDA implementation. In turn, we choose each contingency in the list as the target, with every other contingency as the source. This leads to 21 x 22 pairs of source and target contingencies. 
Fig. \ref{fig:hypothesis3_jda} shows the average balanced accuracy achieved by three methods with JDA and the TFM with few-shot learning. 
\textit{JDA-best} is computed by selecting the best source contingency for every target. \textit{JDA-best} can be computed only post-hoc; that is, we evaluate the best source contingency for each target by comparing the predicted labels from each JDA source model with the actual target labels. Hence, \textit{JDA-best} is the theoretical upper bound on achievable accuracy with JDA-based transfer learning for the set of contingencies considered. Note that \textit{JDA-best} cannot be deployed in practice because it requires (unavailable) full labels from the target contingency to choose the best source contingency for the transfer learning.  Meanwhile, \textit{JDA-EDC} selects the source contingency for every target by choosing the electrically closest contingency from the training database. This is done by computing the \textit{L2} distance between the EDC vector of the target contingency and every possible source contingency in the training database, and choosing the source with the smallest distance. Similarly, \textit{JDA-mod-EDC} selects the source contingency for each target by choosing the closest one based on the proposed mod-EDC vector. With \textit{TFM+10}, we use TFM with mod-EDC encoding and add 10 additional labelled samples from the target domain to the model's context.
\textit{JDA-best} outperforms both \textit{JDA-EDC} and \textit{JDA-mod-EDC} as we would expect, since both EDC and mod-EDC cannot correctly choose the best source for every target without additional labelled data from the target domain. Fig. \ref{fig:heatmap_jda} shows the mutual transfer accuracy (MTA) between contingency pairs. We observe that MTA shows a strong correlation with the label distribution, $P(y)$. 
Two contingencies with similar label distributions have high MTA, and vice versa. Surprisingly, \textit{TFM+10} performs comparably with \textit{JDA-best} on held-out contingencies, using just 10 additional labelled samples randomly generated from the OC distribution, as described above. Unlike typical transfer learning, where the model must be fine-tuned with additional weight-update steps, \textit{TFM+10} requires no additional weight updates. The additional labelled samples can simply be added to the model context, and the model automatically takes them into account during prediction. Furthermore, Tab. \ref{tab:JDA_results} compares the four approaches across all metrics and highlights TFM's strong generalization performance via few-shot learning. TFM+10 matches the oracle JDA-best on 3 of the four metrics, even on the more difficult held-out bus scenario.

\begin{table}[b]
\centering
\caption{Three JDA approaches compared with the TFM+10 approach, which uses few-shot learning with 10 additional labelled samples (mean $\pm$ std). TFM+10 has comparable performance to the oracle JDA-best approach in the held-out contingency generalization scenario. It outperforms the practical JDA with EDC and mod-EDC approaches in the two held-out scenarios.}
\label{tab:JDA_results}
\setlength{\tabcolsep}{3pt}
\footnotesize
\begin{tabular}{c c c c c}
\hline
\textbf{Metric}    & \textbf{Bal. Acc.}     & \textbf{Macro F1}   &  \textbf{Spec.}   & \textbf{Prec.}\\
\hline
\hline
\multicolumn{5}{l}{\textit{Held-out contingency}} \\
JDA (best)   & 87.7 $\pm$ 0.4 & 82.4 $\pm$ 0.5 & 84.8 $\pm$ 0.4 & 85.2 $\pm$ 0.3 \\
JDA (EDC)    & 82.7 $\pm$ 0.3 & 78.6 $\pm$ 0.3 & 78.1 $\pm$ 0.5 & 85.3 $\pm$ 0.1 \\
JDA (mod-EDC)& 77.4 $\pm$ 0.4 & 73.9 $\pm$ 0.5 & 72.6 $\pm$ 0.5 & 79.0 $\pm$ 0.2 \\
TFM+10       & 87.3 $\pm$ 1.7 & 87.0 $\pm$ 1.2 & 83.6 $\pm$ 2.3 & 91.7 $\pm$ 0.8 \\
\hline
\multicolumn{5}{l}{\textit{Held-out bus}} \\
JDA (best)   & 85.2 $\pm$ 0.5 & 78.5 $\pm$ 0.6 & 79.6 $\pm$ 0.4 & 78.1 $\pm$ 0.4 \\
JDA (EDC)    & 74.1 $\pm$ 0.2 & 64.9 $\pm$ 0.2 & 72.0 $\pm$ 0.4 & 76.0 $\pm$ 0.2 \\
JDA (mod-EDC)& 71.1 $\pm$ 0.5 & 63.6 $\pm$ 0.5 & 66.8 $\pm$ 0.6 & 73.3 $\pm$ 0.1 \\
TFM+10       & 80.0 $\pm$ 0.7 & 78.7 $\pm$ 0.7 & 78.2 $\pm$ 2.5 & 82.9 $\pm$ 4.4 \\
\hline
\end{tabular}
\end{table}
\subsection{Computational performance}
ML-based DSA approaches are motivated by the need for near real-time security assessment. Therefore, it is important to analyze the computational cost of the proposed approach. For our evaluation, we consider the hyperparameter tuning, training and inference times separately. The proposed approach has no hyperparameter tuning time. TFM also requires no training; however, it requires time to fit the context samples into memory before inference. We have classified the context fitting time as 'training' time to keep it comparable with the baseline approaches. The computational cost of the TFM mainly consists of the inference time. Tab. \ref{tab:computational_cost} shows the computational cost of the various models. The baseline models incur most of their cost in the hyperparameter tuning step, which scales unfavourably with the training data size and the number of models/contingencies to train. On the other hand, TFM performs ICL with no hyperparameter tuning so tuning cost is 0,however, inference is more computationally expensive than the baseline models. Inference cost also increases sublinearly with the number of training samples in the model context, whereas the baseline models have nearly constant inference time once training is done, except for SVM-RBF.
In the multi-contingency DSA setting, a single TFM requires about 20s to fit the context into memory and afterwards takes about 72s to predict labels for all 12000 test OCs of each contingency. The MDSA-S approach takes about 7s for training and tuning. Afterwards, inference is very fast, taking less than 1s to predict all test OCs for each contingency, but it has inferior predictive performance. The performance improvement the TFM approach offers over the baseline may be considered a reasonable trade-off for the increased computational time. We note that we ran the TFM computation on a GPU, whereas the classical ML baselines can only use a CPU. Also, TFM's computational cost grows significantly with the number of features because of the underlying transformer architecture. In larger systems, this may be challenging. Dimensionality reduction techniques and training data sampling may be used to keep memory consumption and computational costs tractable.

\begin{table*}[htbp]
  \centering
  \caption{Average hyperparameter tuning, training and inference time in seconds per contingency (mean $\pm$ std) for TFM and the baselines. For TFM, training time is the time to fit context into memory since no weight update is done. Inference is done on 12000 test OCs per contingency. TFM, FT-Transformer and GATv2 run on an NVIDIA H100 GPU, while the other models run on a CPU.}
  \label{tab:computational_cost}
  \footnotesize
  \begin{tabular}{|l|r|r|r|r|r|}
    \hline
    \multirow{2}{*}{Model} & \multicolumn{5}{c|}{Training data size} \\
    \cline{2-6}
     & 12 & 120 & 1200 & 12000 & $1200 \times 22$ \\
    \hline
    \multicolumn{6}{|l|}{\textit{Hyperparameter tuning}} \\
    \hline
    TFM            & 0 & 0 & 0 & 0 & 0 \\
    XGBoost        & 0.55 $\pm$ 0.10 & 2.00 $\pm$ 0.16 & 13.69 $\pm$ 1.80 & 51.47 $\pm$ 5.42 & --- \\
    MLP            & --- & 0.62 $\pm$ 0.04 & 3.14 $\pm$ 0.20 & 36.10 $\pm$ 1.49 & --- \\
    DT             & 0.038 $\pm$ 0.005 & 0.14 $\pm$ 0.00 & 2.47 $\pm$ 0.08 & 38.02 $\pm$ 1.81 & --- \\
    SVM-RBF        & 0.063 $\pm$ 0.008 & 0.16 $\pm$ 0.00 & 7.63 $\pm$ 0.95 & 716.4 $\pm$ 100.2 & --- \\
    FT-Transformer & 3.85 $\pm$ 0.44 & 3.18 $\pm$ 0.48 & 34.21 $\pm$ 6.71 & 340.2 $\pm$ 68.5 & --- \\
    GATv2          & 4.81 $\pm$ 0.18 & 3.32 $\pm$ 0.13 & 12.25 $\pm$ 0.51 & 155.4 $\pm$ 4.1 & --- \\
    MDSA-S         & --- & --- & --- & --- & 7.02 $\pm$ 0.03 \\
    \hline
    \multicolumn{6}{|l|}{\textit{Training}} \\
    \hline
    TFM            & 0.29 $\pm$ 0.00 & 0.33 $\pm$ 0.00 & 1.83 $\pm$ 0.10 & 10.80 $\pm$ 0.01 & 19.79 $\pm$ 0.33 \\
    XGBoost        & 0.017 $\pm$ 0.009 & 0.088 $\pm$ 0.008 & 0.55 $\pm$ 0.07 & 2.18 $\pm$ 0.23 & --- \\
    MLP            & --- & 0.020 $\pm$ 0.002 & 0.15 $\pm$ 0.03 & 1.49 $\pm$ 0.24 & --- \\
    DT             & 0.001 $\pm$ 0.000 & 0.007 $\pm$ 0.000 & 0.13 $\pm$ 0.00 & 2.27 $\pm$ 0.19 & --- \\
    SVM-RBF        & 0.002 $\pm$ 0.000 & 0.005 $\pm$ 0.000 & 0.18 $\pm$ 0.03 & 8.55 $\pm$ 0.37 & --- \\
    FT-Transformer & 0.28 $\pm$ 0.08 & 0.15 $\pm$ 0.02 & 1.81 $\pm$ 0.33 & 16.31 $\pm$ 0.49 & --- \\
    GATv2          & 0.33 $\pm$ 0.08 & 0.11 $\pm$ 0.01 & 0.60 $\pm$ 0.03 & 7.73 $\pm$ 0.65 & --- \\
    MDSA-S         & --- & --- & --- & --- & 0.060 $\pm$ 0.001 \\
    \hline
    \multicolumn{6}{|l|}{\textit{Inference}} \\
    \hline
    TFM            & 5.20 $\pm$ 0.01 & 6.69 $\pm$ 0.01 & 10.23 $\pm$ 0.03 & 27.72 $\pm$ 0.03 & 72.27 $\pm$ 0.02 \\
    XGBoost        & 0.005 $\pm$ 0.001 & 0.006 $\pm$ 0.000 & 0.010 $\pm$ 0.001 & 0.019 $\pm$ 0.001 & --- \\
    MLP            & --- & 0.014 $\pm$ 0.001 & 0.014 $\pm$ 0.001 & 0.013 $\pm$ 0.001 & --- \\
    DT             & 0.001 $\pm$ 0.000 & 0.001 $\pm$ 0.000 & 0.002 $\pm$ 0.000 & 0.003 $\pm$ 0.000 & --- \\
    SVM-RBF        & 0.031 $\pm$ 0.001 & 0.11 $\pm$ 0.01 & 0.50 $\pm$ 0.09 & 1.95 $\pm$ 0.15 & --- \\
    FT-Transformer & 0.18 $\pm$ 0.02 & 0.17 $\pm$ 0.02 & 0.17 $\pm$ 0.02 & 0.16 $\pm$ 0.02 & --- \\
    GATv2          & 0.059 $\pm$ 0.004 & 0.053 $\pm$ 0.006 & 0.054 $\pm$ 0.004 & 0.058 $\pm$ 0.004 & --- \\
    MDSA-S         & --- & --- & --- & --- & 0.002 $\pm$ 0.000 \\
    \hline
  \end{tabular}
\end{table*}

\section{Discussions}
This work demonstrates the application of a tabular foundation model to data-driven pre-fault DSA. The proposed TFM approach for pre-fault DSA is promising, outperforming baselines from the literature while demonstrating the potential to reduce the computational cost of generating labelled data. The proposed workflow is also flexible, adapting to single-contingency or multi-contingency DSA settings without performance degradation. Compared with the baselines, the proposed approach is also robust to severe class imbalance. For unseen contingencies, the TFM's sample efficiency becomes a critical operational advantage, as relatively few labelled samples are needed to generalize to new contingencies. Where additional training data for the new contingency cannot be generated, the proposed TFM approach may be combined with EDC encoding of contingency location for zero-shot generalization. The mod-EDC encoding is designed for scenarios where a few additional labelled samples of the new contingency may be generated via TDS. This scenario is operationally relevant, since generating a small number of samples in the online phase is realistic and requires no additional model training.
Overall, the proposed workflow positively answers the three questions posed in Section IV. Our approach offers a viable solution to large database requirements and inadequate generalization to new contingencies. Apart from the strong predictive performance of the proposed approach, TFM also provides additional capabilities relevant to DSA applications. Such capabilities include synthetic data generation for data augmentation, probability density estimation for uncertainty quantification and interpretability analysis for feature selection. We leave the exploration of these aspects for future studies.
Despite these advantages, we note a few key limitations. First, we have assumed a fixed clearing time for all contingencies in this work. Further work is needed to include variable clearing time in the scenario vector. Second, we have used a single test system (IEEE 68-bus) and fault type (3-phase short-circuit) for this study. Third, the computational analysis compares GPU-based deep learning models to CPU-based classical models. Furthermore, separate TFM classifiers may be needed in a DSA workflow which considers many contingencies, for computational efficiency. The TFM we have adopted in this work is limited to inputs of 100,000 samples and 2000 features, which may be inadequate for large systems. Inference time for the TFM is currently slower than lightweight ML classifiers, especially for large systems. As with other ML-based pre-fault DSA approaches, our approach does not guarantee perfect stability prediction; therefore, system operators should validate the model output before making a decision.

\section{Conclusion}
This paper studies the use of a pretrained TFM for pre-fault DSA. The TFM works by ICL and needs no retraining or hyperparameter tuning. Through case studies on the IEEE-68 bus system with 22 contingencies, we demonstrate the TFM's sample efficiency and accuracy. First, the TFM achieves an average Macro F1 score of about 90\% with only 120 labelled samples per contingency. Labelling 120 OCs for each contingency via TDS only takes about 24 CPU seconds compared to about 40 CPU minutes for 12000 OCs, a typical training dataset size in the literature. Second, the TFM is robust for contingencies where insecure cases are rare. In a contingency with only 1\% of its samples insecure, TFM attains an insecure class specificity of about 59\%, compared with the multi-label DSA baseline, MDSA-S, which reaches only about 3\%. Third, the proposed modified EDC encoding enables strong generalization to unseen contingencies via few-shot learning. For held-out contingencies, TFM with modified EDC encoding and 10 additional labelled targets added to its context attains an average balanced accuracy of 87\%, while the best possible transfer learning approach (JDA-best) reaches 88\%. On new contingencies from held-out buses, TFM attains a balanced accuracy of 80\%, while JDA-best reaches 85\%. JDA-best is an oracle model and is not deployable in practice. In the future, we will extend this approach to a larger test system, such as the French transmission system. We will use PCA and other dimensionality reduction techniques to improve scalability, and we will assess the model’s interpretability with partial dependence plots to improve operator confidence in its predictions.

\section*{Declaration of Generative AI Use}
Some parts of the manuscript benefited from grammar and clarity improvements using a generative AI tool, namely Claude by Anthropic. The authors remain responsible for all content.

\bibliographystyle{IEEEtran}

\bibliography{references_edited}


 





\end{document}